\documentclass{article}
\usepackage{ijcai21}

\usepackage{times}
\usepackage{soul}
\usepackage{url}
\usepackage[hidelinks]{hyperref}
\usepackage[utf8]{inputenc}
\usepackage[small]{caption}
\usepackage{graphicx}
\usepackage{amsmath}
\usepackage{amsthm}
\usepackage{booktabs}
\usepackage{algorithm}
\usepackage{algorithmic}
\title{Attention Models for Point Clouds in Deep Learning: A Survey}

\author{
Xu Wang$^1$
\and
Yi Jin$^1$\and
Yigang Cen$^1$\and
Tao Wang$^1$\And
Yidong Li$^1$
\affiliations
$^1$Beijing Jiaotong University\\

\emails
\{xu.wang, yjin, ygcen, twang, ydli\}@bjtu.edu.cn,
}

\begin{document}

\maketitle

\begin{abstract}
  Recently, the advancement of 3D point clouds in deep learning has attracted intensive research in different application domains such as computer vision and robotic tasks. However, creating feature representation of robust, discriminative from unordered and irregular point clouds is challenging. In this paper, our ultimate goal is to provide a comprehensive overview of the point clouds feature representation which uses attention models. More than 75+ key contributions in the recent three years are summarized in this survey, including the 3D objective detection, 3D semantic segmentation, 3D pose estimation, point clouds completion etc. We provide a detailed characterization (1) the role of attention mechanisms, (2) the usability of attention models into different tasks, (3) the development trend of key technology.
\end{abstract}

\section{Introduction}

Point clouds  representation is an important data format that can preserve the original geometric information in 3D space without any discretization. Meanwhile, deep learning have widely and successfully applied to various tasks nowadays. Therefore, it is natural that more and more research currently aims at the adaption of deep learning to 3D point clouds, such as computer vision \cite{wang2019deep} and robotics \cite{behl2019pointflownet}. However, unordered and irregular 3D point clouds structure are still a significant challenge for deep learning. The traditional point cloud representation methods include BEV \cite{yang2018pixor}, multi-view \cite{yang2019learning}, and 3D voxels \cite{maturana2015voxnet}. The main problem of these methods is the fast growth of point sets size \cite{hu2020randla} and geometric information loss \cite{qin2019pointdan}. To alleviate these problems, attention mechanism is introduced to make neural networks to focus on the important parts of input data, helping to simplified point clouds and capture sufficient feature representations \cite{chaudhari2019attentive}. Thus, in this paper, we aim to provide a brief, yet comprehensive survey on attention models for point clouds in deep learning.

There have been a few domain-specific surveys published \cite{nguyen2018attentive,lee2019attention,chaudhari2019attentive,wu2020deep,liu2019deep}. Compared with existing surveys, the contributions of our work can be summarized as:
\begin{enumerate}
    \item To the best of our knowledge, this is the first survey fully focus on attention models for point clouds tasks in deep learning, including computer vision, robotics and miscellaneous applications.
		\item This paper comprehensively covers recent and advanced progresses of attention models for point clouds. Therefore, it allows readers to learn about the state-of-the-art attention mechanisms from different perspectives.
\end{enumerate}

The structure of this paper is as follows. We start by introducing short overview of attention mechanisms in section 2. Then we provide and discuss the attention mechanisms in different tasks in section3 to 5, respectively. In section 6, we present the development trend for future technology. Finally, we conclude the paper in section 7.

\section{Overview}

Human brain system can focus on just a salient regions with limited data, instead of an entire scene \cite{marblestone2016toward}. An attention-based feature extraction is used in the salient regions to acquire the high-level feature representation for improving brain efficiently learns. In spired by this prior knowledge, attention mechanism was first introduced in deep learning to help to researchers choose important data for their tasks \cite{bahdanau2015neural}. With the continuous research, attention mechanisms have achieved great success in Natural Language Processing \cite{9194070}, Computer Vision \cite{wang2016survey}, and robotics \cite{6784518}. Next, we will describe two main attetion mechanism types used in citation papers.

\subsection{Sequence}

{\it Self-attention} means that learning relevant tokens in a single input sequence for every token in the same input sequence \cite{chaudhari2019attentive}. Obviously self-attention is sometimes called inner-attention. Example of self-attention is \cite{yang2019modeling}.

{\it Co-attention}, on the other hand, means that multiple input sequences are processed simultaneously and jointly learn their attentive feature weights to capture interactions between these inputs \cite{chaudhari2019attentive}. Example of co-attention is \cite{you2018pvnet}.

\subsection{Positions}

{\it Soft-attention}is a differentiable deterministic process so that it can be trained with a backpropagation algorithm \cite{kingkan2019point}. The {\it global-attention} model is similar to the soft attention model \cite{luong2015effective}.

{\it Hard-attention}, on the other hand, is a non-differentiable stochastic process and relies on a sampling-based method for training \cite{kingkan2019point}. {\it local-attention} model is an intermediate between soft-attention and hard-attention \cite{luong2015effective}.

\section{3D Computer Vision Applications}

In this section, we review existing attention models for computer vision. We group the applications to different subcategories, namely, 3D recognition and retrieval, 3D detection, 3D segmentation, 3D classification and 3D registration.

\subsection{3D Recognition and Retrieval}

\subsubsection{3D Recognition}

3D object recognition is one of the most fundamental and intriguing problems in computer vision, spanning broad applications from environment understanding to self-driving. Attention model is used to make the neural network focus on informative features to obtain a stronger representation. To gesture recognition, \cite{kingkan2019point} design an automatic feature extraction network by using soft-attention module. This work is based on the intuition that, only particular points of body movement in an input point clouds are required for the network to classify the gestures. \cite{li2019graph} propose a graph attention module to specify different weights for different nodes by calculating relational degree in the local feature space. In their model, attention module is applied four times in different layers to aggregate feature and dynamically update the state of node. Finally, they can obtain more plentiful node feature representation. Similarly, \cite{xia2020soe} apply self-attention unit to better capture feature dependencies among long-range context. \cite{sun2020dagc} introduce a dual attention module (point-wise and channel-wise) to weigh importance of points and features for enhancing the feature representation ability.

Similar to the human visual system, attention mechanism can be used to multimodal feature fusion. \cite{you2018pvnet} propose a point-view feature fusion method based on soft-attention mask. \cite{lu2020pic} use global channel attention and spatial attention VLAD \cite{jegou2010aggregating} to fuse the feature of point cloud and image. \cite{zhao2020manet} present a MANet framework for high-precision 3D object recognition that is able to fuse point-view data. \cite{luo2020learning} introduce an embedding attention point-slice fusion strategy for new shape representation.

\subsubsection{3D Retrieval}

In order to manage large scale point cloud datasets, exploring effective 3D shape retrieval algorithms is necessary. \cite{li2020mpan} propose a multi-part attention network for 3D model retrieval, and applies a novel self-attention module to explore the spatial relevance of local features. \cite{zhang2019pcan} apply a Point Contextual Attention network to discriminate the local feature which positively contribute to the final global feature representations. \cite{lei2019deep} report that view differences of feature have no direct impact on retrieval performance. Their Representative-View Selection algorithm only trend to choose views which can contribute to better performance. In other works, \cite{liu2019l2g} develop a hierarchical self-attention to highlight informative elements in point, scale and region levels. \cite{dovrat2019learning} extend visual attention, focusing the subsequent task network on significant points. Experiments on various benchmark datasets show that these methods can effectively remove the redundancy and results in an enhanced feature representation.

\subsection{3D Detection}

3D object detection is an important aspect in computer vision. However, point clouds are usually unordered, sparse and unevenly distributed, which heavily affects feature extraction and accurate object localization. \cite{paigwar2019attentional} extend visual attention mechanism for multiple object detection. The attention module makes the network focus on smaller region containing the objects of interest. \cite{wu2020realtime} exploit self-attention mechanism to boost useful features and suppress useless features. \cite{xie2020mlcvnet} design two attention modules and a feature fusion module for 3D object detection that are able to exploit contextual information at patch, object and global scene levels. Similarly, \cite{liu2020tanet} propose a Triple Attention module that considering the channel-wise, point-wise and voxel-wise attention jointly. \cite{li2020grnet} present an end-to-end geometric relation network architecture inspired by the self-attention mechanism. To solve boundary problem, \cite{wang2020centernet3d} introduce an auxiliary corner attention module. Its key contribution is to enforce network focus on object boundaries.

\subsection{3D Segmentation}

3D semantic instance segmentation is a popular topic in computer vision. However, there are still many challenges for 3D point clouds segmentation, such as large scene and heterogeneous anisotropic distribution. \cite{qingyong2019randla-net} combine Local Spatial Encoding and Attentive Pooling modules to automatically learn important local feature. \cite{zhang2020attan} propose an Attention Adversarial Network based on adversarial learning. In the learning phase, network can pay more attention to different regional informative features. \cite{tu2020reconstruction} present an online attention-base spatial and temporal feature fusion method for high-precision and real-time semantic segmentation. 4D point clouds (3D point cloud videos) segmentation is a more challenging task, which needs to capture both spatial and temporal information. To solve above problems, \cite{shi2020spsequencenet} design a cross-frame global attention module. Instance segmentation aims to understand geometric information of point clouds on both semantic level and instance level. To instance segmentation, \cite{liang20193d} introduce a graph neural network based on attention mechanism which can aggregate geometric and embedding information from neighbours. \cite{wen2020cf} model the relationships between neighbor and central points by learnable attention mechanism.

\subsection{3D Classification}

3D classification is a critical task in computer vision, which is widely utilized in autonomous vehicle and robotics. In recent years, weak representation ability of low-dimensional feature and noisy points are still challenging. \cite{fuchs2020se(3)-transformers} introduce a robust SE(3)-Transformer, a variant of the self-attention module for data translation and rotation. \cite{lee2019set} present an attention-base network, Set Transformer, to model interactions among elements in point clouds. \cite{yang2019modeling} propose to use attention layers to capture the relations between point. They also design a parameter-efficient Group Shuffle Attention to decrease voluminous computing consumption of Multi-Head Attention. Airborne Laser Scanning (ALS) classification is a critical application for point clouds. \cite{shajahan2019roof} design a view-based approach for roof classification, based on adding a self-attention network. \cite{bhattacharyya2021self} propose an elevation-attention module, urging network take per-point elevation information into account for better ALS classification.

\subsection{3D Registration}

Point clouds registration is a key problem for computer vision, which aims to estimate the optimal rigid transformation between two or more different point sets. However, point clouds have innumerable unique aspects that can increase the complexity of this problem, such as local sparsity and noisy points. \cite{yew20183dfeat} add an attention layer into their network architecture that better identify 3D local keypoints and descriptors for matching. Inspired by this work, \cite{lu2019deepvcp} develop a novel point weighting layer to learning the saliency of each point in an end-to-end framework. \cite{wang2019deep1} combine attention-base module and pointer generation layer to approximate combinatorial matching. \cite{lu2020rskdd} present an Attentive Point Aggregation module that can be used in keypoints generation by aggregating positions and features of neighbor points. Also, this module outputs an attentive feature map help to estimate saliency uncertainly of each keypoint. \cite{qiaoend} use self-attention and cross attention to enhance structure information and corresponding information for feature aggregation.

\section{Robotic Applications}

In this section, we review a variety of attention mechanisms that can applied to robotic tasks. We group the applications as 3D completion,3D pose estimation and scene flow.

\subsection{3D Completion}

Point cloud completion is a challenging problem in robotic and computer vision. Incomplete point cloud shapes cannot be directly used in practical application due to the limited view angles or occlusion. \cite{zhang2020detail} use attention module to reconstruct and refine the input point clouds, the generated points are more uniformly distributed with fewer outliers and noises. \cite{wen2020point} propose a Skip-Attention Network for point cloud completion. Their proposed model can extract geometric information from local regions of incomplete point clouds to encode complete shape representation at different resolutions. \cite{han2020reconstructing} design a Non-local Attention module that combines multi-resolution shape details and contributive local features for shape completion. \cite{zhang2020multi} add an Attention Unit in their multi-stage network. This unit allocates higher weights for the important points which provide more valuable information for point clouds reconstruction.

\subsection{3D Pose Estimation}

3D pose estimation is widely applied in robotic tasks, such as manipulation, grasping and navigation. The key challenge is to estimate pose by extracting enough features of point clouds to find pose in any environment \cite{yuan2020shrec}. \cite{yang20203dsenet} present a 3D Spatial Attention Region Ensemble Network for real-time 3D hand pose estimation. With the help of spatial attention mechanism, they extract enough local structure features of hand joints. 6D pose estimation is another important branch of pose estimation, including 3D rotation and 3D translation. \cite{song2020pam} propose a Point Attention module to extract powerful feature from point clouds, with geometric attention path and channel attention path. This module makes neural network focus on efficient geometric and channel information to create better feature representations. \cite{du2020dh3d} introduce an attention predictor that effectively utilize multi-level geometric information and channel-wise relations to generate global descriptor. Unlike above single input approaches, multimodal inputs can provide additional feature information. \cite{yuan20206d} apply a graph attention network to effectively fuse the color and depth features. Similarly, \cite{cheng20196d} use attention mechanism to learn discriminative multimodal features from image and point clouds. The difference between two works mentioned above is that they use different network architectures.

\subsection{Scene Flow}

Scene flow is the 3D displacement vector between each surface point in two consecutive frames \cite{wu2019pointpwc}. Estimating scene flow is an important fundamental basis for numerous higher-level challenges such as robotics. It is noteworthy that each point in the point clouds has only one direction flowing to the second frame, not all feature information has the same importance. \cite{wang2020hierarchical} propose a hierarchical attention learning network model for scene flow estimation. This model includes two different attention modules, first attentive embedding and second attentive embedding, which can better focus on matched regions and features to find the right flowing direction. Inspired by above work, \cite{wang2020pwclo} present an attention cost volume structure to associate two point clouds and extract the embedding motion information. \cite{puy2020flot} propose FloT attention module for scene flow estimation by optimal transport tools.

\section{Miscellaneous Applications}

In this section, we review the attention model unclassified in preceding two categories, namely, 3D upsampling and 3D normal estimation.

\subsection{3D Upsampling}

Point clouds provide a flexible and scalable geometric representation suitable for a variety of applications, but its unordered and irregular structure also needs to be noticed. To alleviate above challenge, upsampling is proposed to acquire dense and uniform point set from raw point clouds. \cite{li2019pu} propose a point cloud upsampling network, namely PU-GAN, to upsample points over patches on object surfaces. PU-GAN uses adversarial network architecture to train a generator module, which can produce a rich and robust point distributions from the latent space. Avoiding the network tend to poor convergence, they introduce a self-attention unit to enhance the feature integration quality. \cite{liu2019l2g} present an unsupervised upsampling method , named L2G-AE, with deep recurrent neural network. They leverage hierarchical self-attention mechanism to help feature aggregation at three levels of point, scale and region. Conversely, \cite{liu2020spu} propose a self-supervised point cloud upsampling model, named SPU-Net, with graph convolution model. They combine the above two models to simultaneously capture context feature information inside and among local regions. \cite{zhao2020pui} develop a upsampling and completing network called PUI-Net. Noticeable, they apply channel attention mechanism to extract discriminative feature from point clouds.

\subsection{3D Normal Estimation}

3D normal estimation is a fundament task for many high-level applications, including 3d reconstruction, tracking, and rendering \cite{liu2019densepoint}. In previous research, traditional normal estimation method, such as Principal Components Analysis, requires manually tuning  hyper-parameters. Recent methods based on deep learning mainly focus on high-quality 3D normal estimation without manually tuning parameters. \cite{wang2020neighbourhood} propose a temperature adjusted multi-head self-attention module, namely TMHSA, which combined with deep neural network. The TMHSA softly fuses per-point weighted feature from different aspects and outputs high-quality feature representations. \cite{matveev2020geometric} present an attention-based neural network model that can improve neighborhood selection of point clouds and effectively incorporate geometric relations between the points. The use of geometric attention module as a means of extracting global feature representation is motivated by the fact that different quantities are defined locally.

\section{Discussions}

In this section, we discuss additional issues and highlight important challenges in future investigation.
\begin{enumerate}
\item {\it New applications}. With the rapid development of 3D sensors and point clouds technologies, new application domains are constantly and gradually emerging, including autonomous driving, virtual reality and smart wear. It would be necessary to explore other attention mechanisms suit for different applications. Indeed, attention mechanism can be used in many applications not limited to the aforementioned domains.

\item {\it Powerful attention model}. Over the past several years, a large number of different neural network architectures have been proposed, such as generative adversarial network, graph neural network and transformer. These architectures are important as they allow deep learning to handle many real-world cases. Moreover, with the rapid increase of network architecture complexity, it is difficult to extract enough feature representation without introducing more computational cost. Therefore, a light-weight attention model with powerful and meaningful feature representation for different network architecture would be an interesting direction for future investigation.

\item {\it Multimodal feature fusion}. Modern application domains, such as autonomous driving, are commonly equipped with multiple sensors e.g., RGB cameras, thermal, starlight, LiDAR and RADAR to provide a more comprehensive understanding of real-world environment. It is important to effectively capture all task-relevant information from multimodal data especially if there are complex structure involved. Therefore, including but not limited to aforementioned point-view feature fuse methods, an interesting direction for future study is looking at attention-based technologies that can be used to effective and simplified multimodal feature extract and fuse.

\item {\it Energy-efficient balance}. In recent years, capturing 3D point clouds is getting easier, so that the scale of point cloud datasets increases gradually. The majority of methods that calculate an attention-based work may have trouble in scaling effectively to larger point sets. Furthermore, an emerging trend of deep learning is applied to handheld devices. Power consumption, computational cost and memory footprint will be the most significant obstacle. Therefore, it would be useful to explore other ways of applying attention mechanism not simply to boost model accuracy but to balance the model energy-efficient.
\end{enumerate}

\section{Conclusion}

In this paper, we have provided a systematic review of state-of-the-art attention models for point clouds in deep learning. To the best of our knowledge, this is the first work of this kind. We group existing work to three intuitive taxonomies: computer vision, robotics and Miscellaneous applications. We also list several challenges and opportunities for future investigation in the field of 3D point clouds based on attention mechanism.

\bibliographystyle{named}
\bibliography{ijcai21}

\end{document}